\documentclass[sigconf]{acmart}
\usepackage{xcolor}
\usepackage{multirow}
\AtBeginDocument{%
  \providecommand\BibTeX{{%
    \normalfont B\kern-0.5em{\scshape i\kern-0.25em b}\kern-0.8em\TeX}}}

\acmSubmissionID{901}
\copyrightyear{2023}
\acmYear{2023}
\setcopyright{acmlicensed}
\acmConference[MM '23] {Proceedings of the 31st ACM International Conference on Multimedia}{October 29--November 3, 2023}{Ottawa, ON, Canada.}
\acmBooktitle{Proceedings of the 31st ACM International Conference on Multimedia (MM '23), October 29--November 3, 2023, Ottawa, ON, Canada}
\acmPrice{15.00}
\acmISBN{979-8-4007-0108-5/23/10}
\acmDOI{10.1145/3581783.3611892}

\begin{document}

\title{Uncovering the Unseen: Discover Hidden Intentions by Micro-Behavior Graph Reasoning}

\author{Zhuo Zhou}
\orcid{0000-0003-4620-4378}
\author{Wenxuan Liu}
\orcid{0000-0002-4417-6628}
\authornotemark[1]
\affiliation{%
	\institution{School of Computer Science, Wuhan University}
	\city{Wuhan}
	\state{}
	\country{China}
}
\email{305272@whut.edu.cn,lwxfight@126.com}

% \author{Wenxuan Liu}
% \orcid{0000-0002-4417-6628}
% \affiliation{%
% 	\institution{School of Computer Science, Wuhan University}
% 	\city{Wuhan}
% 	\state{}
% 	\country{China} 
% }
% \email{}

\author{Danni Xu}
\orcid{0000-0001-9482-0111}
\affiliation{%
	\institution{School of Computing, National University of Singapore}
	\city{}
	\state{}
	\country{Singapore}
}
\email{dannixu.gracie@foxmail.com}

\author{Zheng Wang}
\orcid{0000-0003-3846-9157}
\authornote{Corresponding authors: Zheng Wang, Wenxuan Liu}
\affiliation{%
    \institution{National Engineering Research Center for Multimedia Software, School of Computer Science, Wuhan University}
	\city{Wuhan}
	\state{}
	\country{China}
}
\email{wangzwhu@whu.edu.cn}

\author{Jian Zhao}
\orcid{0000-0002-3508-756X}
\affiliation{%
	\institution{Intelligent Game and Decision Laboratory}
	\city{Beijing}
	\state{}
	\country{China}
}
\email{zhaojian90@u.nus.edu}
\renewcommand{\shortauthors}{Zhuo Zhou, Wenxuan Liu, Danni Xu, Zheng Wang, \& Jian Zhao}


\newcommand\blfootnote[1]{% 
\begingroup 
\hspace{-0.6em}
\renewcommand\thefootnote{}\footnote{#1}% 
\addtocounter{footnote}{-1}% 
\endgroup 
}
\begin{abstract}
This paper introduces a new and challenging Hidden Intention Discovery (HID) task. Unlike existing intention recognition tasks, which are based on obvious visual representations to identify common intentions for normal behavior, HID focuses on discovering hidden intentions when humans try to hide their intentions for abnormal behavior. HID presents a unique challenge in that hidden intentions lack the obvious visual representations to distinguish them from normal intentions. Fortunately, from a sociological and psychological perspective, we find that the difference between hidden and normal intentions can be reasoned from multiple micro-behaviors, such as gaze, attention, and facial expressions. 
Therefore, we first discover the relationship between micro-behavior and hidden intentions and use graph structure to reason about hidden intentions. To facilitate research in the field of HID, we also constructed a seminal dataset containing a hidden intention annotation of a typical theft scenario for HID.
Extensive experiments show that the proposed network improves performance on the HID task by 9.9\% over the state-of-the-art method SBP.
\end{abstract}




%%\keywords{datasets, neural networks, gaze detection, text tagging}

\maketitle

\section{Introduction}
\begin{figure}[!t]
  \includegraphics[width=\columnwidth]{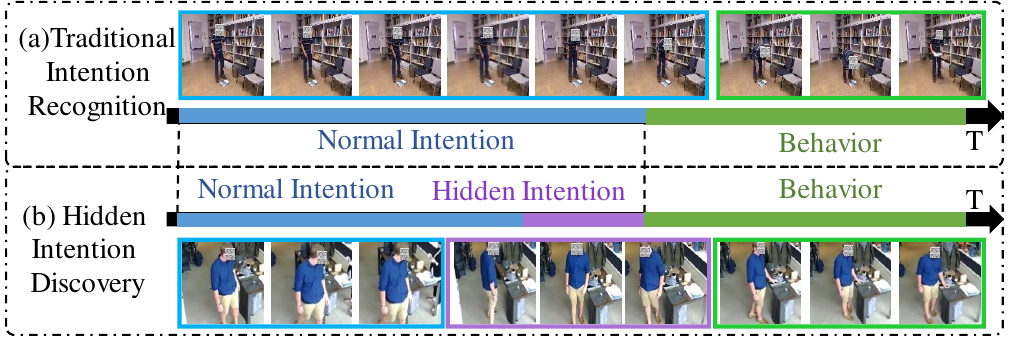}
  \caption{Difference between traditional intention recognition and hidden intention discovery. Take pick-up things and theft alerts as an example, (a) only recognizes normal intention, (b) aims to discover the hidden intention (purple part) which hides in normal intention.}
  \label{fig1}
  \vspace{-1.5em}
\end{figure}
Intention recognition has already been widely explored and applied in Human–Computer Interaction. Previous studies have primarily focused on using surveillance videos to learn visible human intentions. However, many intentions represent fewer visual characteristics for direct machine learning recognition. \textit{E.g.}, the visual difference is often avoided by the actors when they have intentions for abnormal behaviors. The intention for some behaviors that actors intend to hide is referred as hidden intention in this paper. Hidden intentions often precede abnormal behaviors and activities threatening social safety and property. Therefore, hidden intention discovery has significant practical significance. \textit{E.g.}, hidden intention for criminal behaviors can avoid and prevent crimes in advance. In this paper, we investigate a new task of discovering people's hidden intentions when they try to hide what they prepare to commit, named Hidden Intention Discovery (\textbf{HID}). 
This task is challenging in that hidden intentions are often deliberately disguised and are not as obvious as their visible counterparts.

In traditional studies~\cite{FangWRWSLFH20, LiuACLSGN20, ChengXXCLLX22, Feichtenhofer0M19, Royka2022PeopleIC}, the intention is usually equated with normal intention --- the open, obvious and easily observable intention for the normal behaviors, and the task of normal intention recognition is for recognizing this intention and predict the resultant normal behaviors, as shown in Fig.\ref{fig1}(a). E3D-LSTM~\cite{wang2019eidetic} learns obvious spatio-temporal visual representations to predict human intentions. EM-base~\cite{WeiXZZ17} and HAO~\cite{WeiLSZZ18} infer human intentions by combining visual representations and attention. These methods can infer normal intentions from obvious visual representations. 
% danni: 1. 这里有点缺少支撑，建议~概括几个具有代表性的意图检测工作的内容、方法和效果
 
However, HID is a highly complex and challenging task compared to traditional intent recognition tasks, which focus on hidden intentions. As shown in Fig.~\ref{fig1} (b), when hidden intentions are intentionally concealed, they may not have clear or explicit visual representations, making it extremely difficult to distinguish from normal intentions directly through visual information. Therefore, traditional approaches that rely solely on explicit motions may not be effective in HID. The key technical challenge of \textbf{HID} is \textit{how to capture consistent and intention-relevant information under different scenes instead of visual representations which drift with scenes}.

Fortunately, recent research shows that micro-behaviors are effective and observable features that can be used to discover hidden intentions. Micro-behaviors mean tiny, unconscious behaviors of humans, including facial expressions, tone of voice, postures, and other signals. The original word ``micro inequality'' was created by Mary Rowe for social behavior science to refer ``small events which are often ephemeral and hard-to-prove, events which are covert, often unintentional, frequently unrecognized by the perpetrator, which occur wherever people are perceived to be 'different'.''~\cite{rowe2008micro,du2020fight}. In recent years, some micro-behaviors have also been verified as having a close relationship human intention in psychological~\cite{Rolfs2022CouplingPT,NogueiraCampos2019AnticipatoryPA} and sociological~\cite{delaRosa2016VisualAD,Gigliotti2020TheCE} studies, such as gaze direction~\cite{DomnguezZamora2018AdaptiveGS}, attention allocation~\cite{ChongWRR20}, and emotional expressions~\cite{ZhaoCXXL0KPSXYF18,Gndem2022TheNB,VaianiQCG22}. 

We are motivated by these discoveries in psychology, sociology and previous explorations in micro-behaviors video cooperation and we introduce the value of micro-behaviors in the field of intention recognition. We reintroduce the definition of human micro-behaviors as subconscious or unconscious behaviors like Mary's but for visual detection research. Because these micro-behaviors happen in subconscious situations, \textit{i.e.}, their behavioral expressions are difficult to control consciously. They are critical visual signals for the hidden intention discovery task. We hope the renewed and diverse micro-behaviors can benefit the HID domain.

Furthermore, the need for labeled data for hidden intentions poses another challenge in HID. Unlike traditional intention recognition tasks that often have labeled datasets available for training, hidden intentions are, by definition, intentionally concealed, making it challenging to collect large-scale labeled data for HID. This scarcity of labeled data makes it challenging to train accurate and robust models for HID. Therefore, we construct a new benchmark: Hidden Intention Dataset (HD). Unlike traditional datasets, our dataset contains unique hidden intention categories with few visual representations, and we start with preparing theft as the initial version of this benchmark.

To bridge the significant research gap in HID, we propose a novel approach that leverages micro-behaviors to infer hidden intentions, named Hidden Intention Discovery Network (HIDN). Our approach is based on the understanding that micro-behavior cues, such as gaze patterns or attention shifts, may reveal the intentions of individuals, even when their intentions are intentionally concealed. We design a Micro-Behavior Module and a Hidden Relationship Graph Building Module to discover and build a hidden relationship graph. We propose a novel Hidden Intention Reasoning Module to reason about hidden intentions.

Our main contributions are summarized as follows: 
\begin{itemize}
\item To our knowledge, we are the first to raise the task of Hidden Intention Discovery (\textbf{HID}). Different from existing normal intention recognition tasks, our task is more practical and challenging. We also introduce micro-behaviors from a sociological and psychological perspective and clarify the definition of micro-behavior for HID task. The Micro-behavior has theoretical stability in HID tasks, which provides a new perspective for discovering hidden intentions in different scenarios.

\item We construct a Hidden Intention Dataset (\textbf{HD}) containing surveillance video clips of hidden intention, normal intention and abnormal behavior. For the first time, our research classifies hidden intentions as a distinct intention category. This categorization provides a new benchmark for understanding and analyzing hidden intentions that may not be readily observable. We believe HD dataset will benefit the artificial intelligence communities in human intention's visual perception. 

\item We developed a Hidden Intention Discovery Network (\textbf{HIDN}) that combines visual information with psychological knowledge for discovering hidden Intention from surveillance data. Our experimental results demonstrate the effectiveness of our approach, showcasing significant improvements on our constructed dataset. This research contributes to the advancement of hidden intention discovery and highlights the potential of our proposed method for real-world applications in surveillance and visual recognition.
\end{itemize} 
% \vspace{-1.3em}
\section{Related Work}
\subsection{Intention Recognition}
Recent research~\cite{WeiXZZ17, WeiLSZZ18, FangWRWSLFH20, LiuACLSGN20} has shown successful utilization of automated planning techniques for model-based intention recognition of human actions. These approaches rely on sequences of previously performed actions to generate plans and project possible future actions for recognizing intentions. Additionally, some studies~\cite{XuHXWLL21, XuHWLLZ22} have explored using nonverbal cues, such as gaze, as crucial signals in human nonverbal communication for intention recognition. However, these methods are proposed and evaluated in situations with obvious intent, limiting the prospects of discovering hidden intentions. In contrast to the popularity of human actions for building intention recognition models, hidden intentions have been relatively under-explored.
%danni: 2. 这里的工作需要确定综述得是不是足够全面，可以概括但是最好全面一点

\subsection{Action Recognition}
The existing deep-learning methods for action recognition can be classified into two types. The first is based on the two-stream network~\cite{SimonyanZ14, LinGH19, Zhong2022GrayscaleEC, Feichtenhofer0M19, ZhongZLJJHW22, LiuZZJWL23}, which takes RGB frames and optical flows as input for each stream. Simonyan \textit{et al.}~\cite{SimonyanZ14} first proposed the two-stream ConvNet architecture for action recognition. SlowFast~\cite{Feichtenhofer0M19} involves a slow and a fast pathway that operates at low and high frame rates and explores several variations to capture spatial semantics and motion with precise temporal precision. The second type is based on 3D convolutions neural network~\cite{TranBFTP15, CarreiraZ17, Feichtenhofer20, FayyazRDN0GG21, LiuZJJL22}, which are designed to capture the spatial-temporal feature jointly. The first 3D CNN for action recognition is C3D~\cite{TranBFTP15}, which models the spatial and temporal features together. X3D~\cite{Feichtenhofer20}, a step-wise network expansion approach, reduces the complexity by expanding one axis in each stage. Traditional action recognition relies on visible action representation, and recognition accuracy drops significantly when there are no obvious action features. 

\subsection{Micro-behaviors }
In recent years, the computing research community starts to study diverse micro-behaviors, which do not have unified definitions, including fine-grained interactions in online behaviors \cite{meng2020incorporating} and tiny real-world behaviors captured with sensors \cite{rani2021exploring} and cameras \cite{walls2020ai,ZhaoLLYF20}. Walls and Bradley L \cite{walls2020ai} define micro-behaviors as behaviors at $1/30$th of a second and utilize facial actions to detect liars in videos and achieve good results. TIPS~\cite{ColarussoCGSRJ23} aims to enhance human-computer interaction by observing user micro-behaviors (speaking and nodding) to explore the structure of blocks in memory. Motti~\cite{Motti2015MicroIA} and Rani~\cite{RaniCL21} optimize human-computer interaction by analyzing different micro-behaviors and attitudes of humans towards different wearable devices. 

However, these methods can only observe the micro-behavioral manifestations of people and do not reflect their intentions. Although liar detection is related to intention detection \cite{walls2020ai}, it is limited compared with diverse intentions. The current work has spatial limitations in facial and head regions and has a rigid definition of time-limitation of one-thirtieth of a second. The rigid definition without exploring spatial or temporal relationships missed efficient information for intention discovery. In this work, we did an initial exploration that utilized graph structure to connect four types of micro-behaviors to unveil the hidden intention of theft in videos.

\begin{figure*}[t]
	\centering
	\includegraphics[width= 0.88\textwidth]{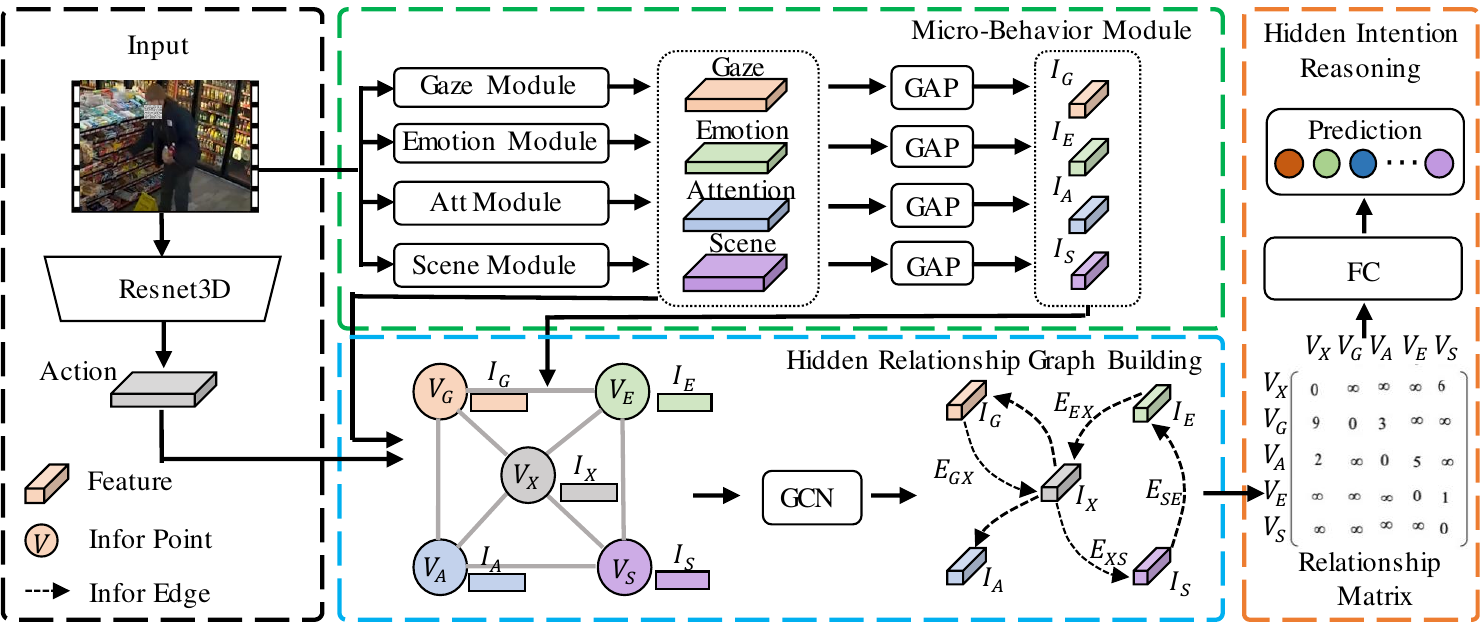} 
	\caption{\small The framework of Hidden Intention Discovery Network. The Micro-Behavior Module is used to learn the gaze, emotion, attention, and scene information. The Hidden Relationship Graph Building module discovers the relationship between different micro-behaviors and builds the hidden relationship graph between them. The Hidden Intention Reasoning module is used to reason about hidden intentions by learning the relationship matrix.}
	\label{fig2}
\end{figure*}
\section{Definitions And Dataset}
The proposed Hidden Intention Dataset (HD) is designed to provide a benchmark for studying hidden intention discovery, where individuals deliberately conceal their intentions or actions. It contains three parts: normal intention, hidden intention, and abnormal behavior. The dataset has a unique challenge as hidden intentions lack obvious behavior features, making it difficult to distinguish them from normal intentions. 

\subsection{Related Definitions}
We made the following classification to distinguish the differences between hidden and normal intentions. We defined two types of intentions: hidden intentions that deliberately hide true intentions, and normal intentions that are expressed truthfully.

\textbf{DEFINITION 1 (Hidden Intention).} Hidden intention refers to an individual's motivation or goal that is intentionally concealed or not immediately apparent from their observable behaviors. Hidden intent usually precedes some deceptive or tactical behavior in which an individual attempts to achieve his or her goals without explicitly revealing his or her true motives to others. The ``readiness potential'' was found by Chun \textit{et al.}~\cite{Soon2008UnconsciousDO} to refer ``We found that decoding of the time decision was possible as early as 5s preceding the motor
decision.'' Therefore, we define the 5s before the behavior as the time when the hidden intent happens.

\textbf{DEFINITION 2 (Normal Intention).} Normal intention refers to an individual's conscious and transparent motivation. The holder of the intention does not act to disguise the intent. This type of intention is normally easily observable in their behaviors. 

\subsection{Dataset Category}

This dataset includes categories of hidden intentions, normal intentions, and abnormal behaviors. The formal definitions are presented as follows:

\textbf{Hidden Intention}: The hidden intention category in the dataset includes the preparatory stages of car theft, burglary, goods theft, and money theft. We select theft preparation as a typical scenario containing hidden intentions because thieves usually have theft intentions before their theft behaviors and try to hide their intentions to avoid detection and punishment. The hidden intention label is determined by seeing if any abnormal behavior follows and whether it covers the intention stage of the abnormal behavior. 

\textbf{Normal Intention}: The normal intention category includes behaviors such as walking, sitting, talking and performing everyday tasks. The individuals in this category should not attempt to conceal their intentions. In this dataset, if there is no subsequent abnormal behavior in a video, this video is labeled as normal intention.

\textbf{Abnormal Behavior}: The abnormal behavior category in the Hidden Intention Dataset (HD) comprises videos that show unusual or deviant actions in the given context. The dataset includes a variety of abnormal behaviors, such as theft, vandalism, fraud, erratic movements, and inappropriate behaviors. This category differs from the hidden intention category in that it has actual abnormal behaviors in the videos, while in the hidden intention category, these behaviors are not shown up.

\subsection{Dataset Construction}
HD dataset used in this study is collected from various online sources, including publicly available videos from the web and other datasets. The dataset construction includes data sources, data collection settings, and data annotation procedures.

\textbf{Data Collection Settings}: The dataset is collected from a diverse range of online sources and other datasets, such as YouTube, UCF-Crime~\cite{SultaniCS18}, N-UCLA~\cite{WangNXWZ14}, and other publicly accessible online platforms. These sources provide various video clips capturing real-world scenes, including indoor and outdoor environments, public spaces, retail stores, public transportation, and other real-world locations. In HD dataset, each video clip lasts about 10 seconds. The dataset contains 933 videos involving more than 500 people, which are approximately 140 minutes long. We follow UCF-Crime and select 70\% as the training and 30\% as the test set.

\textbf{Challenges and Considerations}: The data collection and annotation process for hidden intentions posed several challenges and considerations. One major challenge is the subjective nature of identifying hidden intentions, as these behaviors often lack obvious cues and may require inferential reasoning based on contextual information. Hidden intent should be searched 5s prior to abnormal behavior. The annotators need to exercise careful judgment and expertise to accurately identify hidden intentions while considering potential biases or contextual factors. Another challenge is the heavy task of data annotation. We organized 10 annotators who each spent 10 hours annotating 200 videos.

\textbf{Reliability and Validity of the Dataset}: Measures are taken to ensure the reliability and validity of the dataset. The annotators are carefully selected and trained to ensure their expertise and consistency in annotating hidden intentions. Additionally, efforts are made to minimize potential biases in the data sources by collecting videos from diverse online platforms and locations. To ensure data quality and integrity, thorough data cleaning and validation procedures are implemented. Any inconsistencies or discrepancies in the annotations are resolved through discussions and consensus among the annotators, such as seeking suggestions from experts and organizing online meetings to vote on the decision. Additionally, data quality checks are performed to identify and rectify any errors or inconsistencies in the dataset. 
\vspace{-1em}
\subsection{Privacy Statement}
In response to the human privacy involved in this paper, many efforts have been made to address the issue. They are: 1) Facial information will only be shared with academic researchers, and public versions will mask facial information for privacy. 2) The dataset only focuses on learning visual representations and does not identify sensitive information such as human identity. 3) In the final version, further efforts will be used to address ethical issues (including signing agreements with participants, doing more detailed blurring of face, \textit{etc.})

\section{The Proposed Method}
Our Hidden Intention Discovery Network (HIDN) aims to discover the hidden intention and psychological state of humans, which explicitly captures reasoning cues. Fig.~\ref{fig2} comprehensively illustrates the composition of the HIDN, which comprises three modules: the Micro-Behavior Module extracts micro-behaviors from the input videos, the Hidden Relationship Graph Building Module aims to construct a graph to maintain the relationship between micro-behaviors, and the Hidden Intention Reasoning Module reasons the hidden intentions by learning a relationship matrix.

\subsection{Micro-Behavior Module}
The Micro-Behavior Module in the proposed network framework is designed to learn micro-behaviors, including gaze, attention, emotion, and scene information. We briefly describe how to learn each type of micro-behavior in the following, and the details are available in the supplementary.

\textbf{Head Cropping Branch.}
The head cropping branch finds and crops the head region of the person from the image. The 'head crop' section of the network first refers to the Retinaface~\cite{ZhaoXXYF20} to find the position of the head from each frame, crop the head area according to the bounding box and resize the image to 256*256. The cropped head image is available for the network to learn information.

\textbf{Gaze Information.} 
We define the range of gaze as micro-behavior named $I_G$. We calculate the head center point $P \left(x_p, y_p\right)$, the corresponding gaze target point $Q \left(x_q, y_q\right)$, and their representations in terms of camera coordinates $\overrightarrow{P} = \left(x_p^t, y_p^t, z_p^t\right)$ and $\overrightarrow{Q} = \left(x_q^t, y_q^t, z_q^t\right)$. We can obtain the gaze direction $\overrightarrow{D} = \left(d_x, d_y, d_z\right)$, and calculate the range of gaze $\theta_t = \arccos \frac{\left(\overrightarrow{D_t} \cdot \overrightarrow{D_{t+1}}\right)}{|\overrightarrow{D_t}|\cdot |\overrightarrow{D_{t+1}}|}$. Followed by GazeFollow360~\cite{LiSGZZG21}, if $\theta<90^{\circ}$ represents that people are focusing on what is in front of them and are not observing their surroundings. Details are available in the supplementary.

\textbf{Attention Information.}
We define the target of gaze as attention denoted by $I_A$. Unlike gaze information, attention information aims to discover what the individual focuses on. Meanwhile, the full-size feature map learned by the ResNet-50~\cite{HeZRS16} network is fed to the attention learning part, consisting of two convolutional layers and a fully connected layer. The modulation is performed by taking elements from the normalized full-size feature map, and then the attention feature map is clipped so that its minimum value is $\geq$ 0 followed by ~\cite{ChongWRR20}. The final attention feature map is overlaid on the input image for attention visualization.

\textbf{Emotion Information.}
The emotion information is defined as $I_E$. Like the ViPER~\cite{VaianiQCG22}, we use the Py-Feat model~\cite{BahreiniNW16} to extract 20 different Face Action Until from the selected frames, and we exploit the logistic regress or pre-trained model instead of the random forest because deprecated. Moreover, the tool provides the scores of 7 emotions (\textit{i.e.}, anger, disgust, fear, happiness, sadness, surprise, and neutral), slightly different from those proposed by the task. Hence, we include the scores above in the FAU feature set.

\textbf{Scene Information.}
We use Places-CNN~\cite{ZhouLXTO14} model for scene recognition and define it as $I_S$, then take out the region of the image containing the scene that will be estimated and extract its most relevant features. The entire image is then inputted and extracted to provide the necessary scene support. The network of this method consists of 8 convolutional layers with 2-dimensional kernels. Then, we use a global average pooling layer to reduce the features of the last convolutional layer. Finally, we add a batch normalization layer after each convolutional layer and a linear unit to get the results. 

\subsection{Hidden Relationship Graph Building}
We hypothesize that the relationship cues among hidden intentions are independent of each micro-behavior display. Therefore, we propose to represent the relationship between the micro-behaviors in each video as a graph to discover the relationship between the hidden intention and the micro-behaviors. 

For each image, the Micro-Behavior Module treats $N$ target micro-behavior vectors $V = \left\{v_1, v_2, \dots, v_N\right\}$ as $N$ node features and calculates the connectivity (edge presence) between a pair of nodes $v_i$ and $v_j$ by the similarity of their features $\left(Sim\left(i, j\right) = {v_i}^T v_j\right)$. Specifically, we choose the $K$ nearest neighbors of each node, and thus the graph topology is made of the learned node features. Then, a GCN layer is employed to update all micro-behavior activation statuses from the produced graph jointly. The $i$-th micro-behavior $\hat{v_i}$ is generated by $v_i$ and its connected nodes as:
\begin{equation}
    \hat{v_i} = \sigma [v_i + g (v_i + g(v_i, \sum_{j=1}^N r(v_j,a_{i,j})))],
    \label{eq4}
\end{equation}
where $\sigma[\cdot]$ is the non-linear activation, $g$ and $r$ represent differentiable functions of GCN layer, and $a_{i,j}$ denotes the connectivity between $v_i$ and $v_j$.

After that, we further extract the edges of micro-behaviors to minimize the hidden relationship graph. We conduct cross-attention between $v_i$ and $v_j$, where $v_i$ and $v_j$ are individually used as queries, while the action representation $X$ is treated as the key and value. This process can be formulated as follows:
\begin{equation}
\begin{aligned}
    R_{v_i,X}, R_{v_j,X} &= \text{CrossAtt}\left(v_i, X\right), \text{CrossAtt}
    \left(v_j, X\right), \\
    \text{CrossAtt} &= \text{Softmax}(\frac{\omega_q v_i\left(\omega_k v_j\right)^T}{\sqrt{d_k}})\omega_v v_j,
    \label{eq5}
\end{aligned}
\end{equation}
where $\omega_q, \omega_k$ and $\omega_v$ are weights, and $d_k$ is a scaling factor equalling the number of the key’s channels. Meanwhile, we also conducts the cross-attention between $R_{v_i,X}$ and $R_{v_j,X}$ to produce the edge between $v_i$ and $v_j$. Finally, we feed $R_{v_i,X}$ and $R_{v_j,X}$ to a GAP layer to obtain multi-dimensional edge feature vectors $e_{i,j}$ and $e_{j,i}$, respectively. Mathematically, this process can be represented as:
\begin{equation}
\begin{aligned}
    e_{i,j} = &\text{GAP}\left(\text{CrossAtt}(R_{v_i,X}, R_{v_j,X})\right), \\ 
    e_{j,i} = & \text{GAP}\left(\text{CrossAtt}(R_{v_j,X}, R_{v_i,X})\right).
    \label{eq6}
\end{aligned}
\end{equation}
Finally, we build the hidden relationship graph $G$ = $(V, E)$ that consists of $N$ node features and $N * N$ multi-dimensional directed edge features.

\begin{table}
	\centering
    \caption{\small Performance comparison with the state-of-the-arts on HD. The best results are shown in bold. $\dag$ indicates reproduced results on HD.}
    \vspace{-1em}
    \small
    \setlength{\tabcolsep}{1mm}
	\begin{tabular}{lccccc} 
	\toprule
	Method & Venue & Precision & Recall & F1 & Accuracy\\
	\midrule
	Two-Stream~\cite{SimonyanZ14}$\dag$ & NeurIPS '14 & 25.3 & 19.5 & 22.0 & 23.7\\
	C3D~\cite{TranBFTP15}$\dag$ & ICCV '15 & 27.5 & 20.4 & 23.4 & 26.3\\
	TSN~\cite{WangXW0LTG16}$\dag$ & ECCV '16 & 30.1 & 21.3 & 24.9& 28.9\\
	I3D~\cite{CarreiraZ17}$\dag$ & CVPR '17 & 31.3 & 22.5 & 26.2 & 30.0\\
	TSM~\cite{LinGH19}$\dag$ & ICCV '19 & 34.6 & 30.3 &32.3 &33.3\\
	SlowFast~\cite{Feichtenhofer0M19}$\dag$ & ICCV '19 & 35.0 & 35.3 &35.1  & 34.2 \\
	X3D~\cite{Feichtenhofer20}$\dag$ & CVPR '20 & 33.3 & 26.9 &29.8 &30.8\\	3DResNet+ATFR~\cite{FayyazRDN0GG21}$\dag$ & CVPR '21 & 37.5 & 38.2 & 37.8 & 36.6\\
	SBP~\cite{ChengXXCLLX22}$\dag$ & CVPR '22 &40.8  & 45.7 & 43.2 & 38.3\\
	\cmidrule{1-6}
	HIDN (Ours) & & \textbf{50.0} & \textbf{66.6} & \textbf{57.1} & \textbf{48.2}\\
	\bottomrule
	\end{tabular}
	\label{table2}
        \vspace{-2em}
\end{table}

\subsection{Hidden Intention Reasoning}
We design a Hidden Intention Reasoning module that computes a relationship matrix to reason about hidden intentions through micro-behaviors and hidden relationship graphs. The multi-dimensional edge features are generated from the last GCN layer to a shared FC layer to recognize the influence of the micro-behaviors in hidden intentions discovery by hidden relationship graph. The relationship matrix (Ma) between each micro-behavior can be obtained from the graph, and the final feature is calculated as follows:
\begin{equation}
\begin{aligned}
    F &=  I \cdot Ma \cdot I^T  \\
    &= [I_X, I_G, I_A, I_E, I_S]  
    \begin{bmatrix}
     &e_{X,X} \dots &e_{X,S} \\
    &\vdots\quad \ddots &\vdots\quad\\
    &e_{S,X} \dots &e_{S,S}\\
    \end{bmatrix}
    \begin{bmatrix}
    I_X\\ I_G\\ I_A\\ I_E\\ I_S 
    \end{bmatrix},
    \label{eq7}
\end{aligned}
\end{equation}
where $I$ is the micro-behaviors and $e$ represents the relationship edges with different micro-behaviors, $F$ denotes the feature of the final reasoning. As a result, the categorical cross-entropy loss is introduced as:
\begin{equation}
\begin{aligned}
    \mathcal{L}_{ce} = -\frac{1}{N} \sum_{i=1}^{N} \left(\hat{y}\ln y + (1-\hat{y})\ln(1-y)\right),
    \label{eq8}
\end{aligned}
\end{equation}
where $y$ represents the prediction result of $F$, and $\hat{y}$ denotes the ground truth.
\begin{table}
	\centering
    \caption{\small Human and model performance. Hidden Intention Discovery Network (HIDN) achieves the best results are shown in bold.}
    \vspace{-1em}
    \small
    \setlength{\tabcolsep}{1.2mm}
	\begin{tabular}{ccccc} 
	\toprule
	Methods & Hidden Intention & Normal Intention  & Average & AHE\\
	\midrule
	Human &32.5 &40.2 &36.4 & 32.5\\
        \midrule
	Baseline &21.9 &25.3 &23.6 & 25.6\\
        SBP &24.2 &32.1  &28.2 & 28.3\\
        \midrule
	HIDN (ours) &\textbf{40.7} &\textbf{43.6} &\textbf{42.2} &\textbf{37.8}\\
	\bottomrule
	\end{tabular}
	\label{table3}
        \vspace{-2em}
\end{table} 
\section{Experiments}
\subsection{Implementation Details}
We use ResNet3D~\cite{TranBFTP15} as the baseline. Starting with the ResNet-50~\cite{HeZRS16} network, a 2D temporal convolution module with a temporal kernel size of 3, followed by batch normalization (BN) and ReLU non-linearity is inserted after every 3D spatial convolution module. A dropout of 0.2 is used between the global pooling and the last fully-connected layer. The initial learning rate is set to 0.01, and decay is set to 0.0001. The network is implemented by PyTorch on one NVIDIA Tesla P100 with a GPU of 12GB memory and trained for 60 epochs.

\subsection{Evaluation Metric}
We evaluated the model in a classification task and a retrieval task, respectively, where the classification task initially evaluated the model's classification effectiveness, while the retrieval task evaluated the model more precisely.
\subsubsection{Classification Metric}
To evaluate classification performance, We follow the HFD~\cite{XuHWLLZ22} and use the Accuracy (commonly used measures in the field of intention recognition), Precision, Recall, and $F1$ score (commonly used measures in the field of information retrieval) as the classification evaluation metrics. 

We define the Average Hidden Effect (AHE), where the predicted scores for hidden $C_H$ and normal $C_\text{Nor}$ actions are obtained by applying the trained model to the samples in the dataset. The AHE metric is then calculated as the average difference between the scores for hidden and normal intentions, according to the following calculation: 
\begin{equation}
\begin{aligned}
    \text{AHE} = \frac{1}{M} \sum_{i=1}^M |C_H^i - C_\text{Nor}^i|.
    \label{eq11}
\end{aligned}
\end{equation}
The AHE metric is an appropriate measure for classification performance on the dataset because it considers the difference between the scores for hidden and normal intentions, which is the critical feature that distinguishes the hidden intention discovery task from traditional intention recognition tasks. All 200 combinations of hidden intention samples are calculated for distance. AHE is calculated by averaging these distances ($i$ now refers to the $i$-th combination, and $M$ represents the number of combinations). A higher AHE score indicates that the model can better distinguish between hidden and normal intentions, which is the ultimate goal of the hidden intention discovery task.
\subsubsection{Retrieval Metric}
To evaluate retrieval performance, we follow the existing retrieval literature~\cite{WangGLLM0PHJS21} and report standard metrics $R@K$ (recall at rank $K$, higher is better). All clips in the original video constitute a gallery. For each original video, the hidden intention clip in the test set is used as a query to retrieve the most similar clip in the gallery at a cosine distance. If the top K video clips with the highest similarity score contain a hidden intention clip, it is considered to be a hit.

\subsection{Human Trials} 
To measure human classification performance on our dataset followed by ~\cite{SmithMY0STU19,ShuBGSLGSTU21}, an experiment in which 10 participants between the ages of 19 and 30 are invited to determine whether hidden intentions would occur in the video. Each participant is assigned to two works containing 20 hidden intention samples. In work 1, participants are asked to determine whether it is a hidden intention and score it from 0 (it is not a hidden intention) to 1 (it is a hidden intention). In work 2, they are asked to determine whether it is a normal intention and score it from 0 (it is not a normal intention) to 1 (it is a normal intention). All responses are filtered through predetermined accuracy and consistency criteria.

\section{Results and Analysis} 
%danni: 标题里面有的有句号，有的没有，需检查一遍~
\subsection{Comparison with the SOTA Methods}
Tab.~\ref{table2} reveals the classification performance of currency methods and Hidden Intention Discovery Network (HIDN) on the Hidden Intention Dataset (HD). For accuracy, HIDN is 9.9\% and 11.6\% higher than the current optimal method SBP~\cite{ChengXXCLLX22} and 3DResNet+ATF~\cite{FayyazRDN0GG21}, respectively, which are mainly based on learning obvious action visual representations, while hidden intentions have no obvious visual representation. HIDN, on the other hand, is designed to learn micro-behaviors and discover hidden intentions, which is crucial for detecting hidden intentions. Using a relation-aware network to build a reliable relationship graph, HIDN can reason about hidden intentions and achieve better accuracy on hidden intention discovery compared to traditional methods.

HIDN outperformed SBP and 3DResNet+ATF by 9.2\% and 12.5\% in accuracy and 13.9\% and 19.3\% in F1, respectively. In particular, our recall metrics improved by 20.9\%, 28.4\% and 39.7\%, respectively, compared with SBP, 3DResNet+ATF and X3D. This indicates that the proposed HIDN method builds a reliable relationship graph that allows for the modeling of the complex relationships between hidden and normal intentions, which enables it to discover hidden intentions. This differs from traditional methods that only learn obvious visual representations of actions. By considering the relationship between hidden and normal intentions, HIDN can better identify hidden intentions indicative of a hidden intention, improving its performance in detecting hidden intentions.
\begin{table}
	\centering
    \caption{\small Recall-at-topK(\%). Hidden intention retrieval experiment of proposed HIDN on HD. The best results are shown in bold.}
    \vspace{-1em}
    \small
    \setlength{\tabcolsep}{3.4mm}
	\begin{tabular}{lcccccc} 
	\toprule
	method  & R@1 & R@3 & R@5 & R@10\\
	\midrule
    Baseline & 5.2  &15.7  & 27.2 & 48.9\\
    SBP & 6.9  &18.5  & 30.4 & 52.7\\
    \midrule
	Baseline \textit{w/} (MBL) &7.6  &19.2 &35.8 & 60.2 \\
	Baseline \textit{w/} (MBL \& HRGB) &8.5 &20.9  &39.7 & 62.9 \\
    \midrule
	HIDN (Ours) &\textbf{10.2}  &\textbf{24.7} &\textbf{43.9} & \textbf{66.5} \\
	\bottomrule
	\end{tabular}
	\label{table4}
        \vspace{-2em}
\end{table} 

\begin{figure}[t] 
	\centering
    \includegraphics[width=0.95\columnwidth]{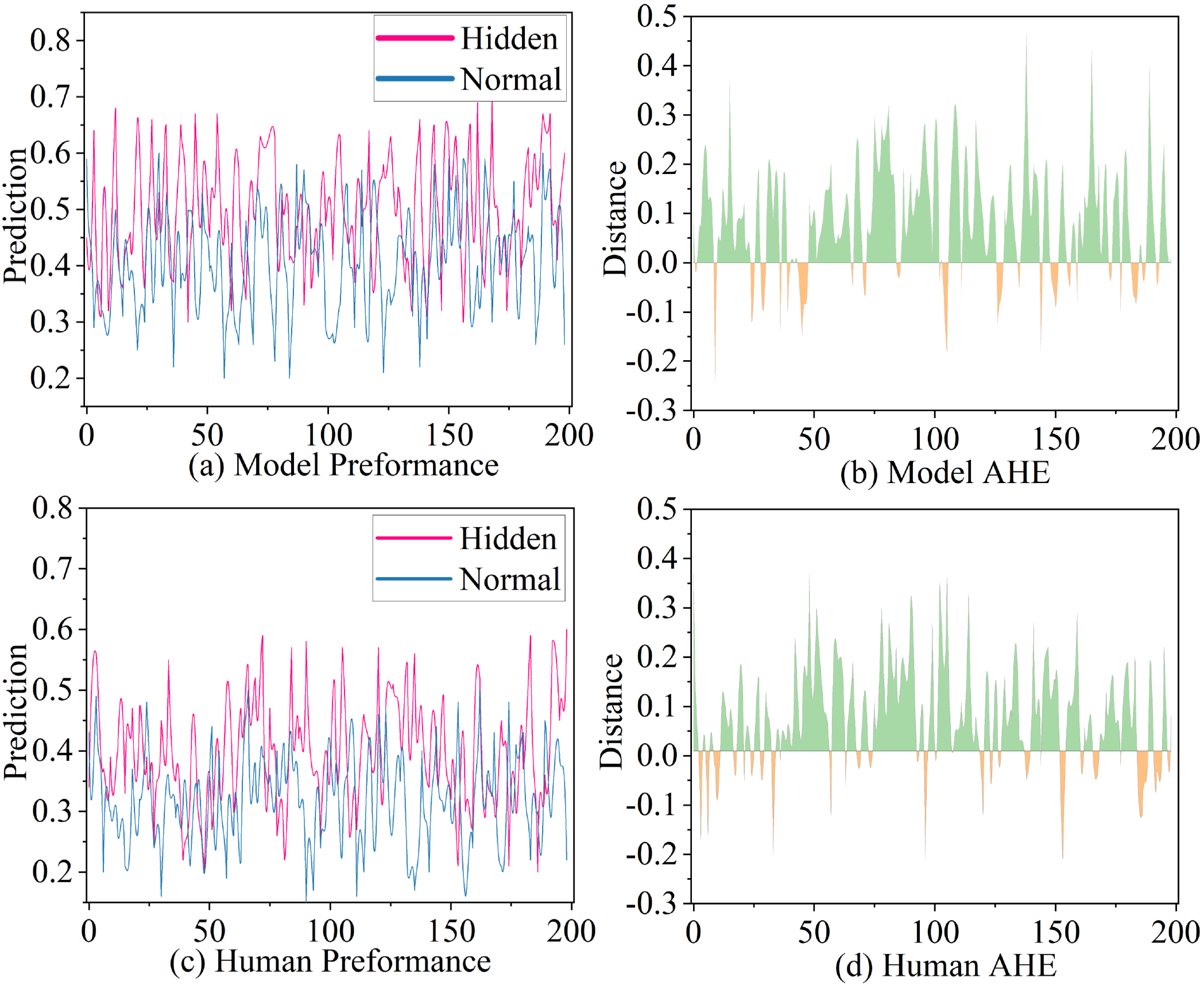} 
	\caption{\small The model Hidden Effect results of the Hidden Intention Dataset. (a) and (c) represent the predicted score of the model and human. (b) and (d) are the distance between the hidden intention score and the normal intention score of the model and human, the green area represents the distance between the hidden intention and the normal intention when the prediction is correct, and the yellow area is the distance when the prediction is wrong.}
     \vspace{-1em}
	\label{fig3}
\end{figure}
\subsection{Human VS Model Performance}
The results presented in Tab.~\ref{table3} and Fig.~\ref{fig3} provide interesting insights into comparing human and model classification performance for hidden intention discovery. Compared to Fig.~\ref{fig3} (a) and (c), the poor performance of humans in hidden intention discovery is a clear indication of the difficulty involved in determining hidden intentions without analyzing the hidden intentions behind them. Moreover, from Fig.~\ref{fig3} (b) and (d), the observation that HIDN is better at distinguishing between hidden intention and abnormal behavior compared to humans, this observation highlights the need for developing effective methods that can analyze micro-behaviors to reason about hidden intentions.

As shown in Tab.~\ref{table3}, the fact that HIDN outperforms both humans and current best methods by a significant margin (8.2\% and 16.5\%, respectively) indicates the effectiveness of the proposed method in discovering hidden intention and reasoning about hidden intentions. This finding corroborates the earlier analysis that HIDN builds a reliable relationship graph for more accurate recognition of hidden intentions. The AHE outperforms humans by 5.3\%, highlighting the difficulty humans face in observing micro-behaviors. The fact that the proposed method can discover micro-behaviors provides an avenue for mining infinite micro-behaviors with limited visual information, thereby making it a promising approach for hidden intention discovery.
\begin{figure}[t] 
	\centering
	\includegraphics[width=\columnwidth]{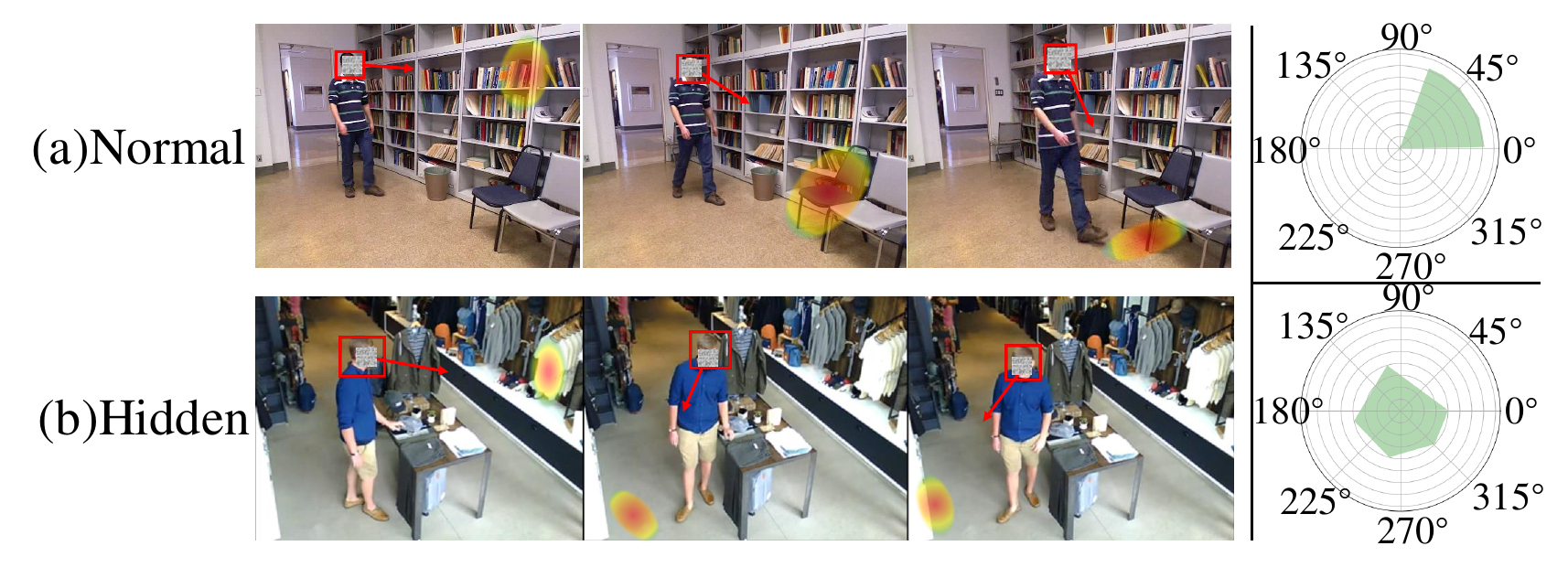} 
	\caption{\small The different gaze attention and the line of gaze between normal intention and hidden intention. The red box and line are the head detection boxes and the gaze direction, respectively. The attention area indicates the area where the person is currently paying attention. The green area on the radar map indicates the range of gaze.}
    \vspace{-1em}
	\label{fig4}
\end{figure}
\begin{table}
	\centering
    \caption{\small Micro-Behaviors analysis experiment of proposed Hidden Intention Discovery Network network on Hidden Intention Dataset. The best results are shown in bold.}
    \vspace{-1em}
    \small
    \setlength{\tabcolsep}{0.6mm}
	\begin{tabular}{lcccccc} 
	\toprule
	Micro-Behavior & Precision & Recall & F1 & Accuracy& Params\\
	\midrule
    Baseline & 35.0 & 35.3 &35.1  & 34.2 &31.6\\
    \midrule
	Baseline \textit{w/} Gaze &42.3 &48.2 &45.0 &39.7 &66.1\\
	Baseline \textit{w/} Attention &39.2 &40.1 &39.6 &36.3 &47.8\\
	Baseline \textit{w/} Scene &37.2 &38.7 &37.9 &35.5 &38.2\\
    Baseline \textit{w/} Emotion &41.4 &41.9 &41.6 &39.5 &68.5\\
    \midrule
    Baseline \textit{w/} (Gaze \& Attention) &47.3 &52.3 &49.7 &45.1 &74.0\\
    Baseline \textit{w/} (Emotion \& Scene) &45.8 &50.9 &48.2 &43.7 &80.8\\
    \midrule
	HIDN (Ours) &\textbf{50.0} &\textbf{66.6} &\textbf{57.1} &\textbf{48.2} &94.5\\
	\bottomrule
	\end{tabular}
	\label{table5}
    \vspace{-1.5em}
\end{table} 
\subsection{Hidden Intention Retrieval}
We also find that localization time is important, so we evaluate the performance of HADN on Hidden Intention Retrieval (finding the start time of the video with hidden intent in the original long video), which is similar to the grounding task to identify the starting and ending time of the video segment described by the query. Following the convention~\cite{BenaimELMFRID20, WangGLLM0PHJS21}, the network is fixed as a feature extractor after pre-training on the Hidden Intention Dataset. Then, the complete original video is divided into clips in 5s units. Tab.~\ref{table4} shows the accuracy at K=1, 3, 5, 10 and compares with baseline and the current optimal method SBP on HD. It can be seen that when using baseline, combining MBL and HRGB modules can bring a 3.3\% improvement for top1 acc and a 5.2\% improvement for top5 acc. In addition, when using the HR module, our results outperform the currently dominant method SBP, which proves that the extracted representations have more substantial discriminative power.

\subsection{Ablation Study}
\subsubsection{Micro-Behaviors Analysis}
The ablation experiments in Tab.~\ref{table5} provide valuable insights into the effectiveness of incorporating multiple sources of micro-behaviors into the HIDN model. The results show that the inclusion of gaze information alone leads to a significant improvement of about 5.4\% in hidden intention discovery compared to the baseline. This finding supports the notion that the gaze can reflect the hidden intentions of humans, and demonstrates the importance of incorporating such information into the model.

Furthermore, adding attention information to gaze leads to a further improvement of about 3.1\% in performance, highlighting the value of attention information in identifying objects of interest. This suggests that multiple sources of micro-behaviors can complement each other, leading to more accurate detection of hidden intentions. Including emotion and scene information further improves the performance by about 9.5\%, indicating that these sources of information can provide additional cues that help identify hidden intentions. This finding suggests that a combination of multiple sources of micro-behaviors can effectively recognize hidden intentions. Future research should investigate the effectiveness of other sources of micro-behaviors in improving performance. 

Moreover, as shown in the Tab.~\ref{table5}, adding gaze and attention increases params by 17.9M and accuracy by 5.4\% compared to adding only gaze. However, adding only attention increases params by 15.8M compared to baseline, but accuracy only increases by 2.1\%, proving that micro-behavioral information improves hidden intent recognition more than params.

\subsubsection{Module Analysis}
The experimental results presented in Tab.~\ref{table6} provide evidence for the effectiveness of each module in the proposed network for hidden intention discovery. The Micro-Behavior Module (MB) has shown to be crucial, achieving a 5.3\% performance improvement over the baseline. This indicates that discovering multiple micro-behaviors is closely related to hidden intentions, and the network can benefit from learning such information. The Hidden Relationship Graph Building Module (HRGB) has also contributed significantly, with a 5.6\% improvement, indicating that building relationships between different micro-behaviors can help discover hidden intentions behind hidden intentions. Finally, adding the Hidden Intention Reasoning module (HR) has improved the performance by 2.1\%, indicating that constructing a reliable relationship matrix is vital in hidden intention discovery. This demonstrates the importance of modeling micro-behaviors and their relationships in uncovering the hidden intentions behind human actions. The findings of this study have implications for the development of intelligent systems that can reason human behavior in complex environments.
\begin{figure}[t] 
	\centering
	\includegraphics[width=\columnwidth]{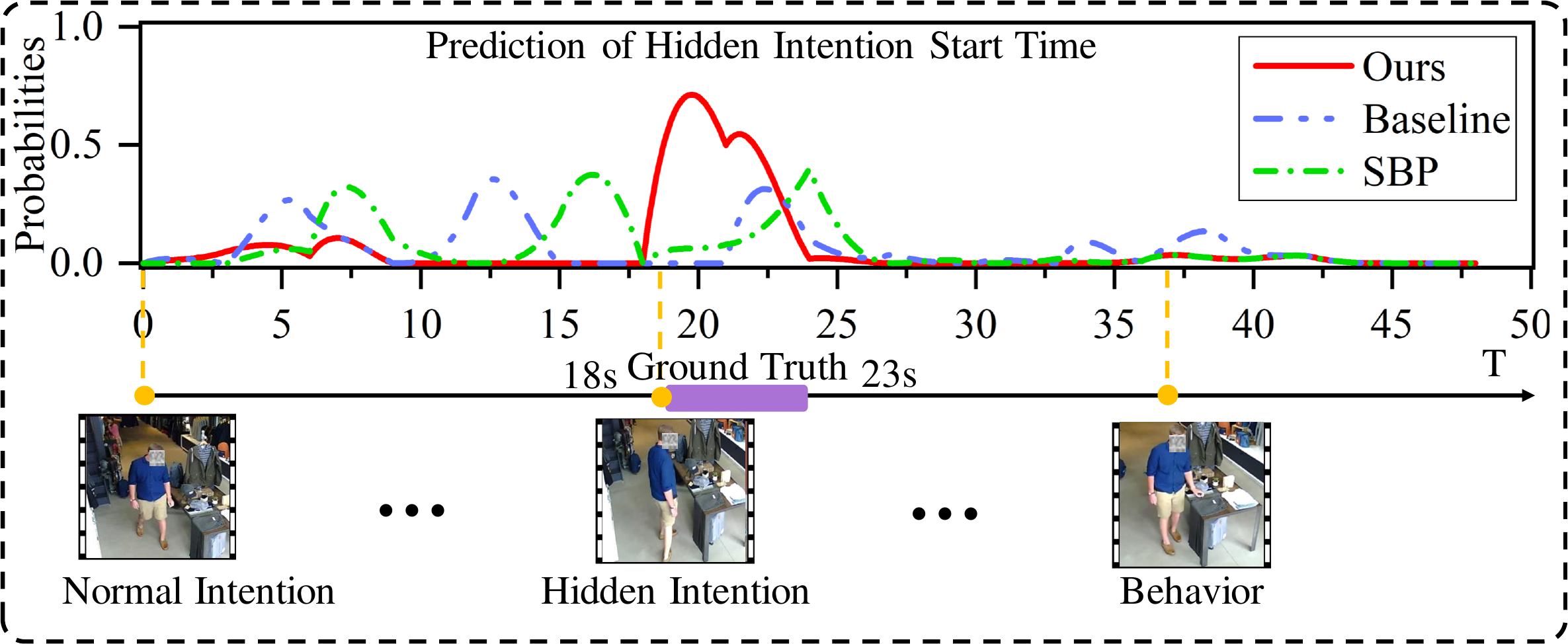} 
    \vspace{-2em}
	\caption{\small The prediction of the hidden intention start time. }
	\label{fig5}
 \vspace{-2em}
\end{figure}
% \subsection{Visualization}
\subsubsection{Gaze Visualization}
Fig.~\ref{fig4} shows the change in the line of gaze between hidden intention (b) and normal intention (a). hidden intention (b) represents the change of the line of gaze in hidden intentions. The thieves observe their surroundings to ensure that the abnormal behavior is undetected. As a result, the line of gaze and radar graph have a wide range of sight. In contrast, in normal intentions (a), people tend to focus only on what is in front of them, with a small range of changes in the line of gaze. This analysis provides strong evidence that the range of the line of gaze can be used as a reliable indicator of hidden intentions, which can be integrated into multimedia-based systems for hidden intention discovery.

\subsubsection{Hidden Intention Retrieval Visualization}
Fig.~\ref{fig5} illustrates the different performances of different methods to localize hidden intentions. Baseline and SBP have predicted normal intention as hidden intention several times before Ground Truth, demonstrating the difficulty of traditional methods to distinguish normal intention from hidden intention in the absence of obvious visual features. However, our method predicts almost nothing before Ground Truth. However, our method predicts almost no errors before ground Truth and can accurately predict when hidden intention occurs due to our method's use of micro-behaviors.

\begin{table}
	\centering
    \caption{\small Module analysis of proposed Hidden Intention Discovery Network on Hidden Intention Dataset. The best results are shown in bold.}
    \vspace{-1.5em}
    \small
    \setlength{\tabcolsep}{4.2mm}
	\begin{tabular}{ccccccccc} 
	\toprule
	Baseline & MBL & HRGB & HR & Accuracy\\
	\midrule
	\checkmark & $\times$ & $\times$ & $\times$ & 34.2\\
	\checkmark & \checkmark & $\times$ & $\times$ & 39.5\\
	\checkmark & \checkmark & \checkmark & $\times$ & 46.1\\
	%\checkmark & \checkmark & $\times$ & \checkmark & 42.3\\
    \checkmark & \checkmark & \checkmark & \checkmark & \textbf{48.2}\\
	\bottomrule
	\end{tabular}%}
    \vspace{-1em}
	\label{table6}
\end{table}
\section{Conclusion}
We highlight an interesting and valuable question: Hidden Intention Discovery (HID). We propose a novel approach for discovering hidden intentions through micro-behavioral feature construction and graph reasoning. We introduce a new dataset, the Hidden Intention Dataset (HD), which consists of videos capturing various scenarios of normal intentions, hidden intentions, and abnormal behaviors. We construct hidden relationship graphs to capture gaze, attention, emotion, and scene information, and use graph reasoning to discover hidden intentions. Our experiments show that the proposed approach achieves superior performance compared to existing methods, demonstrating the potential of micro-behavior analysis in discovering hidden intentions.

\section{Broader Impact}
This work focuses on a new task: Hidden Intention Discovery. The potential broader impacts of this work are listed as follows:

\noindent\textbf{Benefits:} Hidden Intention discovery can benefit many applications, including early detection of crime, prediction of team tactics, and judging human relationships. The study of hidden intention also contributes to the understanding of human behavior and has research value.
Our proposed model relies less on obvious visual features, incorporates sociological and psychological knowledge, and improves the hidden intentions discovery. 

\noindent\textbf{Risks:} As noted, our new dataset has room for further experimentation. This paper focuses on only one type of hidden intentions, and more categories and styles should be an extension. Richer additional support, such as trajectory analysis, may be incorporated in the future. In the long term, more and larger datasets will be needed to improve generalizability. Ultimately, this knowledge will facilitate the development of hidden intention discovery. 

\begin{acks}
This work was supported by National Key R\&D Project (2021YFC3320301), National Natural Science Foundation of China (62171325,62006244), Hubei Key R\&D Project (2022BAA033), CAAI-Huawei MindSpore Open Fund, Young Elite Scientist Sponsorship Program of China Association for Science and Technology (YESS20200140), and Young Elite Scientist Sponsorship Program of Beijing Association for Science and Technology (BYESS2021178).
\end{acks}
\bibliographystyle{ACM-Reference-Format}
\bibliography{sample-base}

\newpage
\appendix
\section*{Appendix}

\section{ADDITIONAL DETAILS OF DATASET}
In order to present our dataset, we provided additional definitions of the Hidden Intention Dataset (HD). As shown in Table~\ref{tablea}, The biggest advantage of HD over other datasets is that it is annotated with hidden intentions, and HD outperforms other datasets in terms of sample size, sample complexity, and behavioral categories.

In addition, we have noticed the following phenomenon in the labeling process. There might be people with hidden intentions who do not conduct abnormal behaviors finally. These people might have been seen as having normal intentions in this dataset and become interruptions. We can call it the type-I error in the dataset. However, this type of mistake cannot be avoided because we have no sound proof to judge people with no subsequent behaviors as hidden intentions. Moreover, it is impossible to directly eliminate "seemingly" Type I errors based on subjective factors due to the risk of losing important normal samples. For this problem, we release a subset of the dataset with hidden intent, normal intent, and type I errors, respectively, and annotated by humans to explore the type-I errors.

\begin{figure*}[t] 
	\centering
        \includegraphics[width=0.93\textwidth]{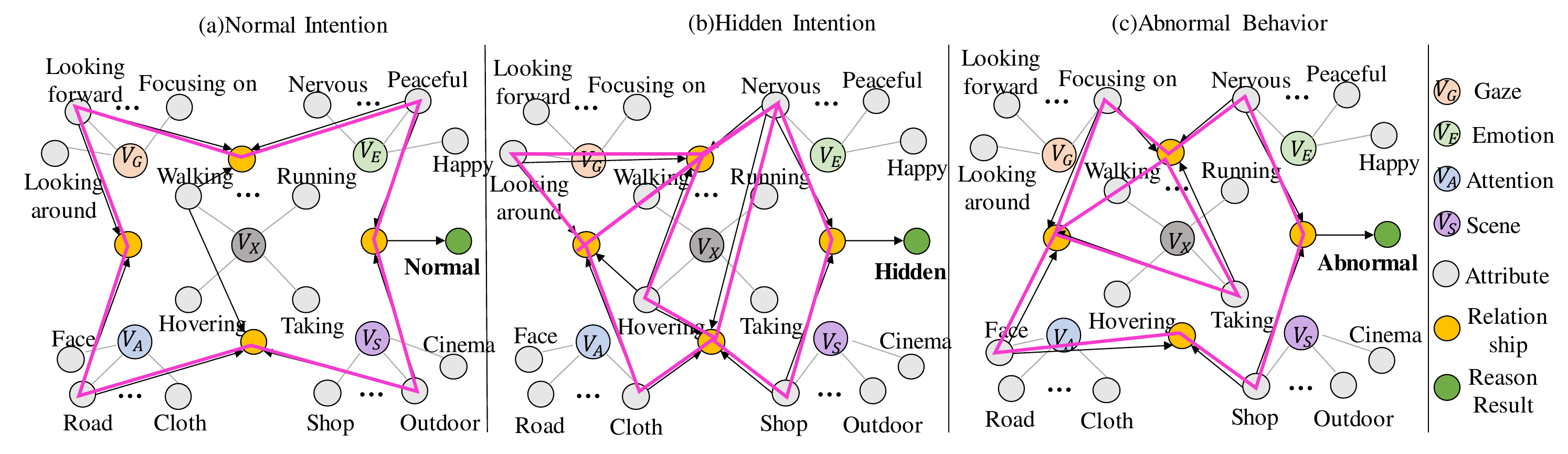} 
	\caption{\small The different hidden relationship graph between hidden intention, normal intention and abnormal behavior. The pink graphs indicate the hidden relationship graph composed of each category. The reason result nodes are the category of action reasoning via the graph structure. The relationship nodes represent the connections between different information. Attribute nodes represent the different attributes of micro-behavior.}
 \vspace{-1.5em}
	\label{figa}
\end{figure*}

\section{Additional Data with Micro-Behaviors Analysis}
In Table~\ref{tableb}, different micro-behavior is added separately to the baseline of the ablation study. When one micro-behavior is added, the best results are obtained by adding gaze. When two hidden messages are added, the best results are obtained by adding Gaze and Attention. The best results are obtained by adding gaze, attention and emotion to the three micro-behavior. The experiments show that including gaze information significantly improved the network's recognition of hidden intentions. The gaze information reflects the thief's observation of the surroundings and shows the thief's hidden intentions.
Similarly, the combined use of gaze and attention information positively influenced the network's recognition of hidden intentions. However, the addition of gaze and scene information did not significantly improve the network's recognition ability compared to the addition of attention. Instead, the combined use of emotion and attention information is less effective than the combined use of emotion and scene information. This suggests that finding the right information relationship is crucial for reasoning about hidden intentions. 

\section{Additional Data with Hidden Relationship Graph}
The graph building analysis shows the complexity involved in reasoning about hidden intentions. Fig.~\ref{figa}(b) shows that the hidden intention relationship graph has the most complex structure because of the need to construct relationships between micro-behavior and reason from multiple aspects. This suggests that discovering micro-behavior and intentions is critical to reason about hidden intentions accurately.

Moreover, the comparison of the normal intention relationship graph (Fig.\ref{figa}(a)) and the abnormal behavior relationship graph (Fig.\ref{figa}(c)) highlights the difference in complexity between normal intentions and hidden intentions. The normal intention relationship graph is relatively simple, while the hidden intentions relationship graph has a more complex structure due to the abnormal behavior of the thieves. This emphasizes the need for methods that can effectively discover and reason about hidden intentions in order to accurately recognize and prevent abnormal behavior.
\begin{table}
    %\vspace{-2.0em}
    \caption{\small Comparison of the proposed Hidden Intention Dataset (HD) with other datasets.}
    \vspace{-1em}
	\centering
    \small
    \setlength{\tabcolsep}{1.5mm}
	\begin{tabular}{lccccccc} 
	\toprule
	Dataset & Hidden Intention  & Quantity & Scene & action categories \\
	\midrule
	UCSD &$\times$ &200 & 2 & 5 \\
	UMN &$\times$ &11 & 2 & 3 \\
        PETS 2009 &$\times$ &72 & 1 & 10 \\
        UH &$\times$ &150 & 8 & 10 \\
        UCF-Crime &$\times$ &1900 & 202 & 13 \\
        \midrule
	\textbf{HD} &\checkmark &933 & 591 & 13 \\
	\bottomrule
	\end{tabular}
    \vspace{-1em}
	\label{tablea}
\end{table}
\begin{table}
	\centering
    \caption{\small Micro-behavior analysis experiment of proposed Hidden Intention Discovery Network on Hidden Intention Dataset. The best results are shown in bold.}
        \vspace{-1em}
	\small
        \setlength{\tabcolsep}{0.1mm}
	\begin{tabular}{lccccc} 
	\toprule
	Micro-Behavior & Precision & Recall & F1 & Accuracy\\
	\midrule
	Baseline \textit{w/} Gaze &42.3 &48.2 &45.0 &39.7 \\
	Baseline \textit{w/} Att &39.2 &40.1 &39.6 &36.3 \\
	Baseline \textit{w/} Scene &37.2 &38.7 &37.9 &35.5 \\
        Baseline \textit{w/} Emotion &41.4 &41.9 &41.6 &39.5 \\
        \midrule
        Baseline \textit{w/} (Gaze \& Att) &47.3 &52.3 &49.7 &45.1 \\
        Baseline \textit{w/} (Gaze \& Emotion) &44.1 &45.7 &44.9 &41.2\\
        Baseline \textit{w/} (Gaze \& Scene) &41.9 &45.2 &43.5 &40.5\\
        Baseline \textit{w/} (Emotion \& Scene) &45.8 &50.9 &48.2 &43.7 \\
        Baseline \textit{w/} (Emotion \& Att) &42.8 &45.7 &44.2 &40.1 \\
        Baseline \textit{w/} (Att \& Scene) &40.6 &43.8 &42.1 &37.2 \\
        \midrule
        Baseline \textit{w/} (Gaze \& Att \& Emotion) &48.5 &59.5 &53.4 &47.5 \\
        Baseline \textit{w/} (Gaze \& Att \& Scene) &46.8 &52.4 &49.4 &46.0 \\
        Baseline \textit{w/} (Gaze \& Emotion \& Scene) &47.9 &56.3 &51.8 &47.3 \\
        Baseline \textit{w/} (Att \& Emotion \& Scene) &47.7 &52.2 &49.8 &45.5 \\
        \midrule
	Ours &\textbf{50.0} &\textbf{66.6} &\textbf{57.1} &\textbf{48.2} \\
	\bottomrule
	\end{tabular}
        \vspace{-1em}
	\label{tableb}
\end{table}

\begin{figure*}[t]
	\centering
	\includegraphics[width=0.9\textwidth]{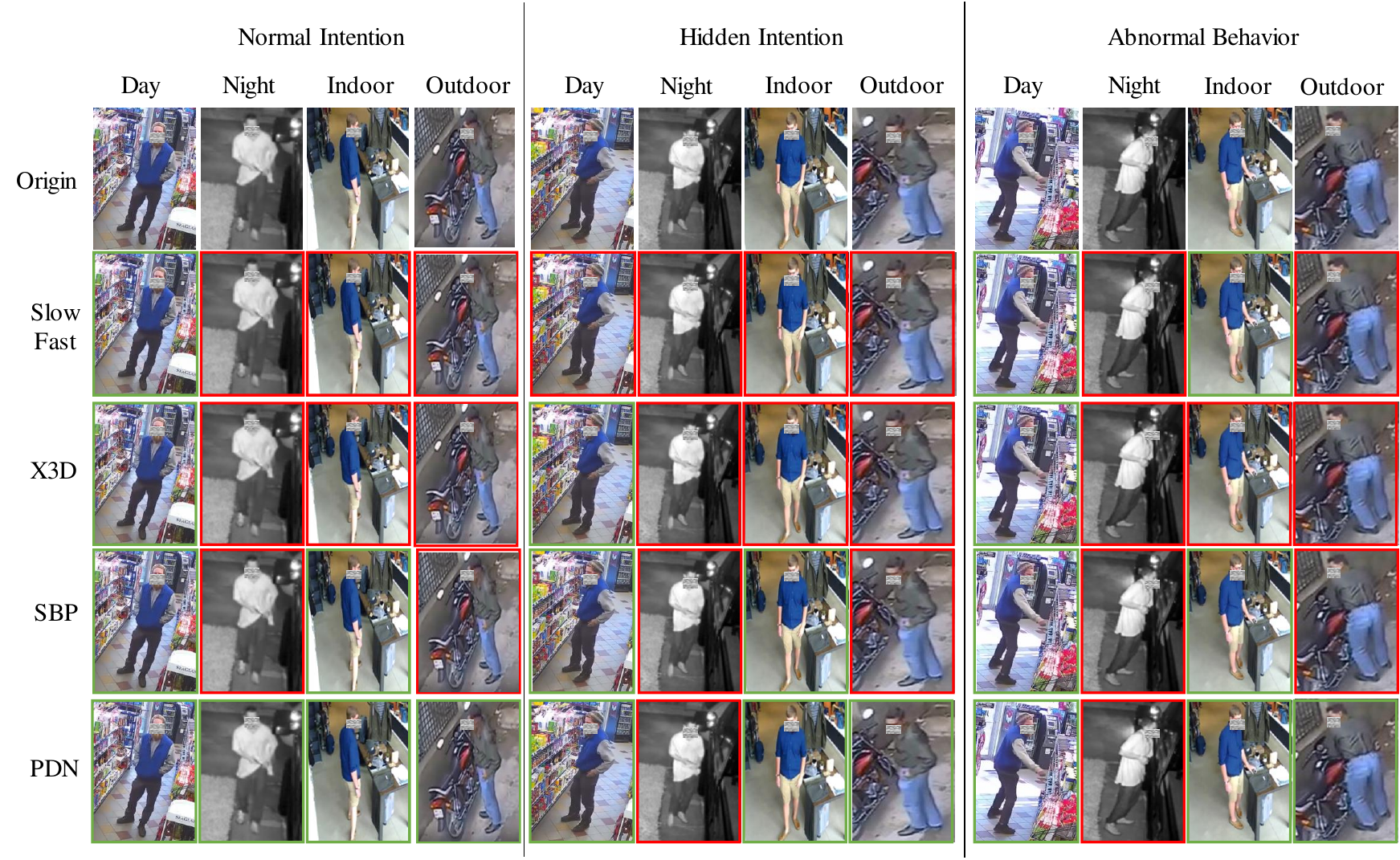} 
	\caption{\small  Comparison of our prediction results and the vanilla methods on the Hidden Intention Dataset. Green boxes indicate correct predictions, and red boxes indicate incorrect predictions.}
	\label{figb}
\end{figure*}
\section{Additional Data with Prediction Results Visualization}
We show our qualitative results for the Hidden Intention Dataset (HD) in Fig.~\ref{figb}. As the results for correct sample prediction show, we observe that several factors can affect the recognition accuracy of the samples, such as self-occlusion, viewpoint changes, lighting changes and camera blur.

We selected some common action categories on the HD dataset and observed the prediction results of these actions using different methods. Normal intention and abnormal behavior in day scenes are more accessible to identify than hidden intention due to the obvious action information for these actions and the positive effect of adequate lighting and video stability on the prediction results.

However, for hidden intention, the discriminative features of these actions are not obvious and traditional methods do not predict these actions correctly because no obvious action information is learned. Our Hidden Intention Discovery Network (HDN) can identify hidden intentions more accurately than other methods by using micro-behavior for a reason. For night scenes, the surveillance camera's erratic sampling blur and dimness make it difficult to learn valid information, resulting in poor predictions. This is an issue that we will focus on in the future.

\end{document}